%% file: acl2023.tex
\pdfoutput=1

\documentclass[11pt]{article}

\usepackage[]{ACL2023}

\usepackage{times}
\usepackage{latexsym}
\usepackage[T1]{fontenc}
\usepackage[utf8]{inputenc}
\usepackage{amsmath}
\usepackage{amssymb}

\usepackage{microtype}

\usepackage{inconsolata}
\usepackage{graphicx}
\usepackage{multirow}
\usepackage{tabularx}
\usepackage{enumitem}
\usepackage{booktabs}
\usepackage{tablefootnote}

\title{Evidence-Grounded Multimodal Misinformation Detection with Attention-Based GNNs} 
 
\author{
  Sharad Duwal$^{1}$\thanks{ \, Correspondence to: \texttt{sharad.duwal@gmail.com}},\,
  Mir Nafis Sharear Shopnil$^{1}$, 
  Abhishek Tyagi$^{2}$,
  Adiba Mahbub Proma$^{2}$ \\ \\
  $^{1}$Fatima Fellowship \, 
  $^{2}$University of Rochester
}

\begin{document}
\maketitle

\begin{abstract}
Multimodal out-of-context (OOC) misinformation is misinformation that repurposes real images with unrelated or misleading captions. Detecting such misinformation is challenging because it requires resolving the context of the claim before checking for misinformation. Many current methods, including LLMs and LVLMs, do not perform this contextualization step. LLMs hallucinate in absence of context or parametric knowledge. In this work, we propose a graph-based method that evaluates the consistency between the image and the caption by constructing two graph representations: an evidence graph, derived from online textual evidence, and a claim graph, from the claim in the caption. Using graph neural networks (GNNs) to encode and compare these representations, our framework then evaluates the truthfulness of image-caption pairs. We create datasets for our graph-based method, evaluate and compare our baseline model against popular LLMs on the misinformation detection task. Our method scores $93.05\%$ detection accuracy on the evaluation set and outperforms the second-best performing method (an LLM) by $2.82\%$, making a case for smaller and task-specific methods.
\end{abstract}

\section{Introduction}

Misinformation has emerged as a major issue with social media \cite{social-media-and-misinformation}. Bad actors disseminate fake information to spread hate, political division, conspiracy theories, health misinformation, and rumors to the disadvantage of targeted groups \cite{wapo-pizzagate,Islam2020COVID19RelatedIA,social-media-and-misinformation}. Visual content is a more viral vector of misinformation than text: fact-checks collected by \citet{dufour2024ammebalargescalesurveydataset} found 80\% of claims contained visual media. Videos are becoming more common in misinformation as of 2022, as are AI-generated media \cite{dufour2024ammebalargescalesurveydataset}. 

\begin{figure}[t]
    \centering
    \includegraphics[width=1\columnwidth]{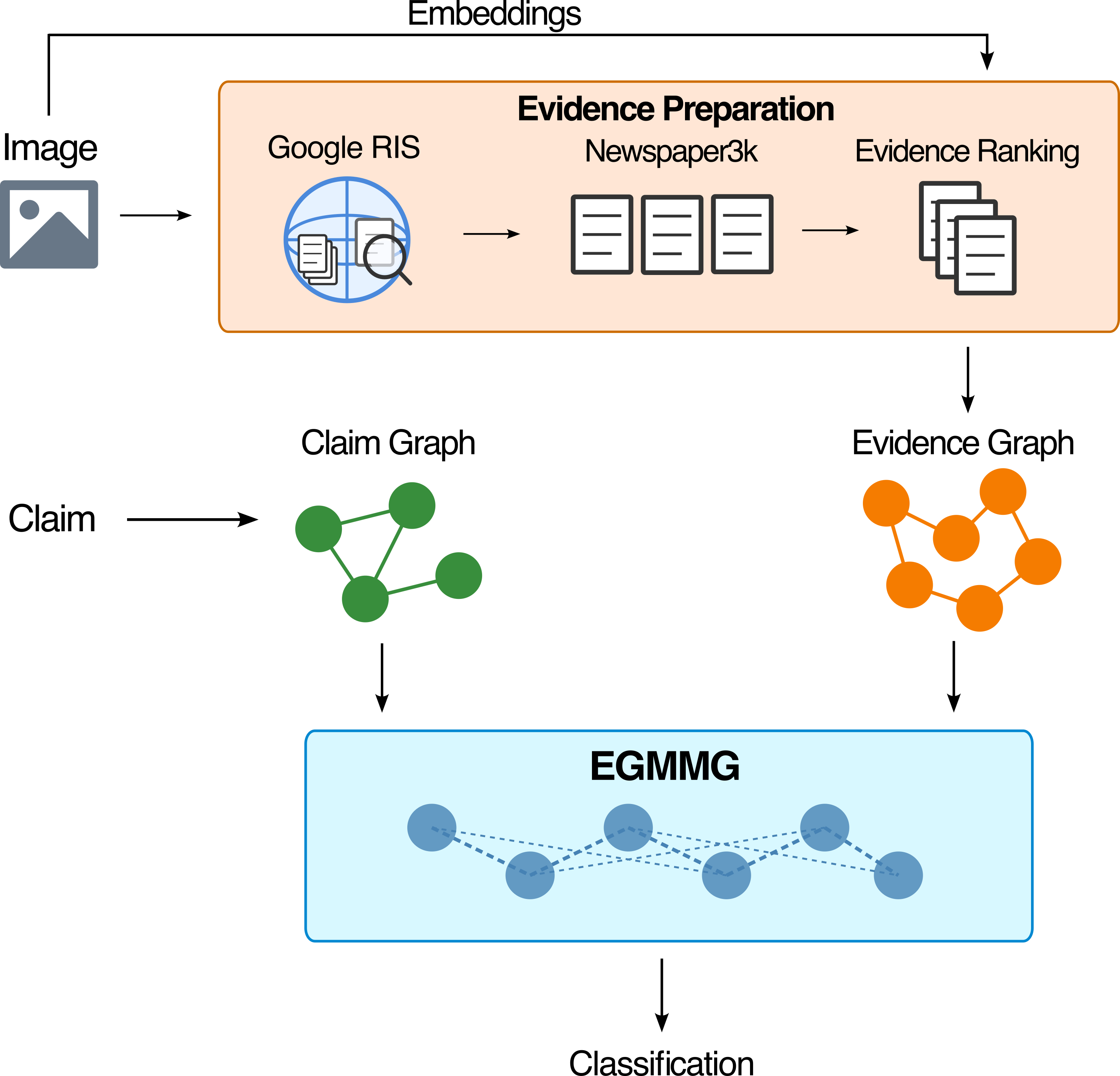}
    \caption{\textbf{The EGMMG pipeline.} For an image-claim sample, the pipeline prepares two graphs, evidence graph and claim graph, using online evidence retrieval followed by a rule-based analysis of subject-object relations in the evidence documents. Once we have the two graphs, we use a graph attention-based classifier to detect misinformation.}
    \label{fig:pipeline}
\end{figure}

\begin{figure*}[t]
    \centering
    \begin{minipage}{\columnwidth}
        \centering
        \includegraphics[width=0.8\columnwidth]{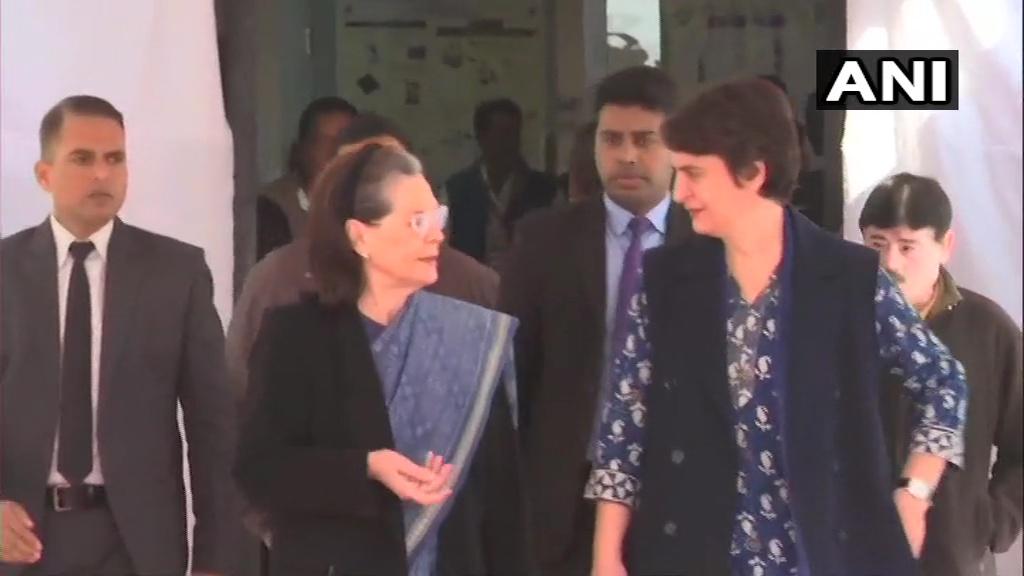}
        \vspace{1mm}
        \caption*{\small \textbf{(a)} Image}
    \end{minipage}
    \begin{minipage}{\columnwidth}
        \centering
        \includegraphics[width=1.1\columnwidth]{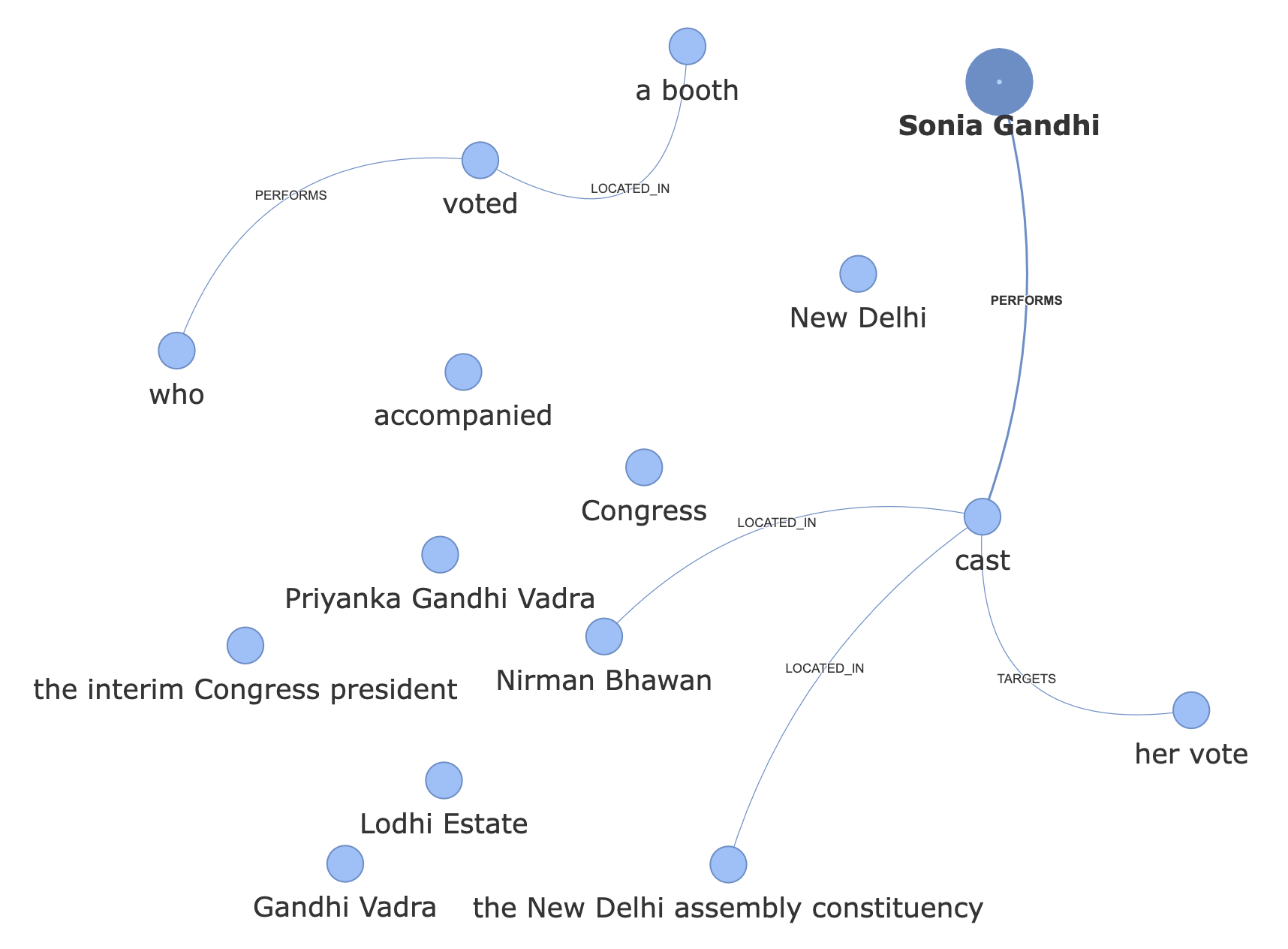}
        \caption*{\small \textbf{(b)} Claim graph generated by EGMMG}
    \end{minipage}%
    
    \vspace{-2mm}
    \caption{\textit{Data sample:} Image and claim graph for claim «\textbf{Sonia Gandhi, the interim Congress president, cast her vote at Nirman Bhawan in the New Delhi assembly constituency, accompanied by Priyanka Gandhi Vadra, who also voted at a booth in Lodhi Estate.}» A section of the evidence graph is provided in the Appendix (Figure \ref{fig:gandhi-voting-comprehensive-evidence}).}
    \vspace{-3mm}
    \label{fig:gandhi-voting-comprehensive}
\end{figure*}

Images repurposed with different captions to make false claims are categorized as out-of-context (OOC) misinformation \cite{fazio2020out}. OOC misinformation is particularly egregious because the image is generally authentic and the misinformation stems from context manipulation \cite{qi2024sniffermultimodallargelanguage}. This makes them more believable, and people, used to photojournalism, tend more readily to accept the claims at face value \cite{fazio2020out}. 

Existing methods to detect OOC misinformation have explored traditional classification algorithms, large language models (LLMs), and large vision language models (LVLMs). While classical methods like feature-based classification were a good starting point, LLMs have proved to be particularly good at detecting (and also explaining) OOC misinformation \cite{qi2024sniffermultimodallargelanguage, xuan2024lemmalvlmenhancedmultimodalmisinformation, tahmasebi2024multimodalmisinformationdetectionusing}. This improvement can be attributed to the world knowledge obtained via pretraining that allows LLMs to be multitask learners \cite{Radford2019LanguageMA, brown2020languagemodelsfewshotlearners}.

An offshoot in OOC multimodal misinformation detection focuses on contextualizing an image claim before reasoning about its truth. Grounding the claim and supporting visual elements using circumstantial evidence assists veracity detection. \citet{tonglet2024imagetellstorypredicting} adopt the ``5 pillars of verification'' as described by \citet{urbani5pils}, which grounds images on five properties: provenance, source, date, location, and motivation to contextualize the images. \citet{tonglet2025covecontextveracityprediction} use these ``pillars'' to establish veracity.

However, a major issue with LLMs is that they ``hallucinate'' when relevant information is not present as parametric information while generating \cite{shuster-etal-2021-retrieval-augmentation, maynez-etal-2020-faithfulness}. Hallucinations are particularly problematic because LLMs can generate explanations that appear credible even when untrue. This is not a desired property in a tool for misinformation detection. 

Retrieval-augmented generation (RAG) \cite{gao2024retrievalaugmentedgenerationlargelanguage}, in-context prompting \cite{brown2020languagemodelsfewshotlearners} and knowledge graphs (KGs) have been introduced to ensure factuality of language models. KGs are effective in adding structured external information; several KG augmented LLMs for misinformation have also been introduced \cite{lu-li-2020-gcan,opsahl2024factfictionimprovingfact, tan-etal-2024-enhancing-fact,wangshu23explainable}. Graph-based methods are widely used in evaluating factuality because they exploit meaningful node relationships \cite{kim2023factkgfactverificationreasoning}.

In this work, we take a graph-based approach to detect multimodal OOC misinformation. We focus on the image contextualization task discussed earlier. We first create an online evidence retrieval pipeline that \textit{hydrates} image-text datasets by collecting textual evidence for the samples using reverse image search (RIS)\footnote{We use Google Vision API.}. The textual evidence found online is used to construct an evidence graph, while the caption in the claim is used to generate a claim graph. We also introduce a baseline graph attention method to learn misinformation detection over the generated graph data.

Our contributions are threefold: 
\begin{enumerate}[nosep]
    \item We introduce a text-grounding approach for the image contextualization task using evidence graphs, which capture the entities in the image and the relations between them,
    \item We introduce a baseline graph attention method to tackle the multimodal OOC misinformation task using the grounding approach,
    \item We pass several publicly available misinformation datasets through the pipeline and establish the model's performance.
\end{enumerate}

\input{relatedworks}

\section{Method}

\textbf{Problem Formalization}\, Given an image $I$ and a textual claim (usually a caption) $C$, the task is to determine a veracity score $s \in [0,1]$ indicating how well the claim supports the image. An image-caption pair will have a high $s$ if the image and the caption are in context (i.e. are related via the subject, object or event).

Misinformation detection that depends solely on images and captions have some issues: i) images might not provide explicit context and ii) captions are often short, single-source, and might also not provide detailed context. These shortcomings present a challenge in establishing veracity.

To tackle this, we focus on the image contextualization task before processing for veracity. We perform reverse image search to obtain resources related to the image with the assumption that these resources (news articles, blog posts, etc.) provide context to the event depicted in the image and claim.

We create an online evidence retrieval pipeline and run it on OOC misinformation datasets to construct contextualized knowledge graph data from the image-caption samples in the datasets. 

We finally introduce a baseline graph attention method to perform misinformation detection on this data.

For our experiments, we focus on positive and negative classes only (for example, \texttt{Refute} and \texttt{Support\_Multimodal} for the Factify dataset). We describe below our evidence retrieval pipeline. Where relevant, we use the Factify dataset 
\cite{Mishra2022FACTIFYAM} as a placeholder, but the pipeline can be easily adapted to other image-caption datasets mentioned in \S\ref{sec:datasets}.

\subsection{Data}
\label{sec:evidence-retrieval}
We start by extracting the claim image, claim text, and the misinformation label from the multimodal misinformation dataset.

For the Factify dataset, there are claim images, support ``document" images, lemmatized claims, lemmatized related document, and a classification label. For our task, we are only interested in the claim image, claim text, and the label. 

To generate the graphs required for our task, we first gather \textit{evidence} for the image (related textual documents on the web) and rank them based on their similarity with the image. Then we use the textual evidence to generate knowledge graphs with entities (subjects and objects) in the text as nodes and relations between them as edges. For an image-caption pair, we follow the steps below to generate the data.

\begin{enumerate}
    \item \textbf{Evidence documents:} We use the Google Vision API to get web pages that use the claim image (full or partial matches). We assume that news articles, essays, and blogs host these images to provide reporting and commentary, which could be useful context. For an image, we try to get at most 30 web pages containing the image. We discard images for which the Vision API does not return at least one web page.
    
    For the web content extraction, we use Newspaper3k. We extract the metadata including text, language, author name, publication date, and time. While the metadata could also be leveraged for detection (similar to \citep{tonglet2024imagetellstorypredicting}), because our focus is on the main text and the entity-relationship, we extract the text content (``evidence'' documents) $E = {e_1, e_2, ..., e_m}$  from web pages containing image $I$.

    As quality check before inclusion as evidence, we rank and filter the web pages. We accomplish this by computing similarity between the embeddings of the web page texts and the image.  We use the \texttt{clip-ViT-L-14} model offered by SentenceTransformers that can embed both images and texts \citep{reimers-2019-sentence-bert}.  We get the embeddings for the web page documents (and page titles) and the image separately and use cosine similarity to get the top-k evidences per image. For evidence document $e_i$, we calculate the similarity score
        \begin{equation*}
            \text{sim}(e_i, I) = \text{cos\_sim}(\phi_I(I), \phi_T(e_i))
        \end{equation*}
    where 
    $\phi_I$ is the image embedding function and 
    $\phi_T$ is the text embedding function.

    After we have the similarity scores for all the evidence documents, we select the top 7 and concatenate them into the final \textit{evidence} for our task.
    
    \item \textbf{Graph Construction:} We construct evidence and claim graphs using the final evidence and the claim texts. Nodes for both the graphs are entities, events, and locations as identified by the \texttt{en\_core\_web\_lg} spaCy model \cite{Honnibal_spaCy_Industrial-strength_Natural_2020}. Relations between the entities are identified based on how they are related: The possible relations (edges) are: \texttt{PERFORMS}, \texttt{EXPERIENCES}, \texttt{TARGETS}, \texttt{LOCATED\_IN}, \texttt{HAS\_STATE}, and \texttt{SAME\_AS}. To annotate these edge types, we use the token POS (\texttt{verb}, etc.) and dependencies (\texttt{nsubjpass}, \texttt{prep}, etc.). More details on graph construction are in Appendix \ref{graph-construction}. We construct the claim graphs similarly using the captions.
\end{enumerate}

\begin{table}[!t]
\centering
\renewcommand{\arraystretch}{1.2}
\begin{tabular}{lrr}
\toprule
\textbf{Dataset} & \multicolumn{1}{c}{\textbf{Orig}} & \multicolumn{1}{c}{\textbf{Final}} \\
\midrule
Factify \cite{Mishra2022FACTIFYAM} & 14000 & 4945 \\
Factify Val& 3000 & 1145 \\
COSMOS \cite{aneja2023cosmos} & 1700 & 813 \\
MMFB Val \cite{liu2024mmfakebench} & 1000 & 391 \\
MMFB Test\tablefootnote{The original set has 10000 samples, but we remove samples with AI-generated images.} & 6750 & 3829 \\
\bottomrule
\end{tabular}
\caption{Dataset statistics showing sample counts before and after evidence retrieval and processing. We do not focus on large train sets due to limited resources.}
\label{tab:dataset_statistics}
\end{table}

\subsection{Model}
We introduce a GNN-based method as baseline for the graph image contextualization task. The method leverages the topological and relational information present in the evidence and claim graphs using cross-graph attention.

The first step is to extract meaningful representations from the created graphs. This can be accomplished by extracting node and edge features.

\subsubsection{Node features}
We use the node label embeddings and node neighborhood information to obtain the node features. We get the label embedding using a language model (BERT \cite{devlin-etal-2019-bert}, for example). For the neighborhood structure, we utilize properties like in-degrees, out-degrees, total degrees, pagerank, and reverse pagerank. 

Since the label's text embeddings are higher-dimensional (depending on the LM of choice; 768 for BERT-base) than the neighborhood structure information (5, for the five structural properties above), we project these two representations to a common dimension during training so that both contribute meaningfully to the node features. For this, we implement a node features projector inside the graph encoder (described in \S\ref{sec:architecture}.) In addition to the linear projections for the label and the structure information, we implement learnable multiplicative coefficients that determine the contribution of each feature to the final node representation.
 
Thus, the embedding of node $v$ at initialization is:
\[
h^{(0)}_v = \alpha \cdot \text{LM}(v) + \beta \cdot \text{NS}(v)
\]
where $LM$ and $NS$ are functions that project language model embedding and neighborhood information respectively into a common space and $\alpha$ and $\beta$ are weight coefficients.

\subsubsection{Edge features}
\label{sec:edge-features}
Since edges inform node relationships, we obtain the edge features as well. The properties we focus on are: edge centrality, common predecessors, common successors, in-jaccard, out-jaccard, forward path length and backward path length.

However, we find that edge features are not beneficial in their current formulation to the architecture on the misinformation detection task (see ablation in Table \ref{tab:ablation-study}). They also add computational overhead while data preparation. For a detailed discussion, please refer to Appendix \ref{graph-construction}.

\subsubsection{Architecture}
\label{sec:architecture}
After initializing the node and edge features for both the evidence and claim graphs as described above, we perform message-passing between the nodes using graph convolutions (GATConv \cite{veličković2018graphattentionnetworks}, TransformerConv \cite{shi2021maskedlabelpredictionunified}, GATv2Conv \cite{brody2022attentivegraphattentionnetworks}) to update the node representations. This allows subgraph neighborhoods to inform each node's representation. The node representation at layer $\ell + 1$ is given by:
\begin{equation*}
    \begin{split}
        h^{(\ell+1)}_v &= \text{GraphConv}\left(h^{(\ell)}_v, \{h^{(\ell)}_u : u \in \mathcal{N}(v)\}\right) \\
    \end{split}
\end{equation*}
where GraphConv is the convolution function, $h^{(\ell)}_v$ and $h^{(\ell)}_u$ are node representations at layer $\ell$ and $\mathcal{N}(v)$ is the neighborhood of the node $v$. 

We experiment with all three convolution methods discussed above and go with TransformerConv for our final model. For experimental results with each convolution type, refer to Table \ref{tab:conv_type_performance_merged} in Appendix.

\begin{figure}[!t]
    \centering
    \includegraphics[width=0.98\columnwidth]{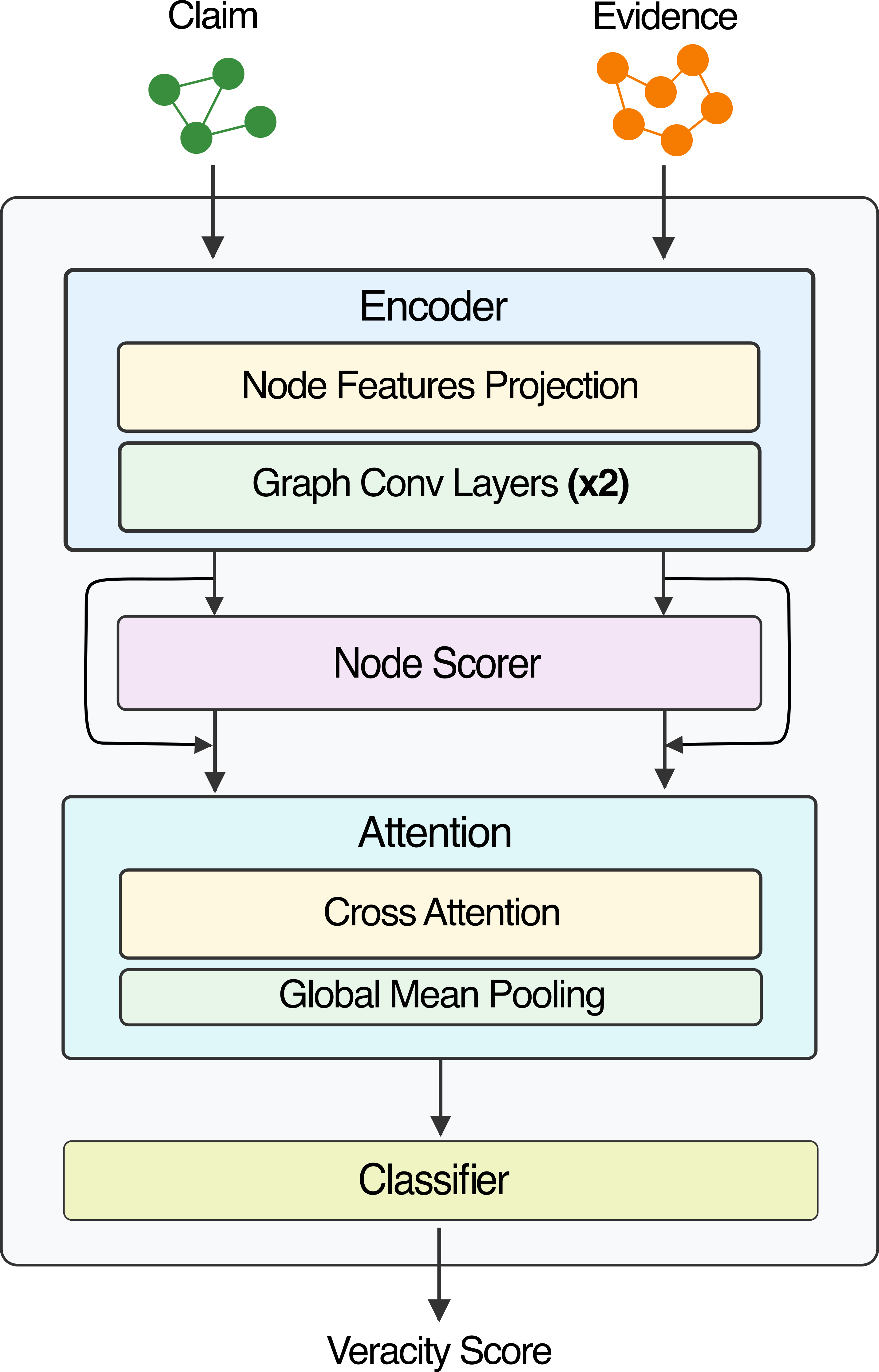}
    \caption{\textbf{The EGMMG classifier.} }
    \label{fig:arch}
\end{figure}

Since all nodes might not be equally important for the detection task, we assign node importance with the help of a trainable node scorer. We multiply the node embeddings generated by the convolution layers with the node scorer, which is a score $s_v \in \mathbb{R}$:

\begin{equation*}
\hat{h}_v = h_v \cdot s_v
\end{equation*}
where $s_v = \text{NodeScorer}(h_v)$ and $\hat{h}_v$ is the importance-weighted node embedding.
 
We then perform cross-attention computation between the evidence and claim graphs. Similar to encoder-decoder cross-attention mechanism in transformers, this gives us the attention pattern between the evidence nodes and claim nodes. We use the claim nodes for the query part of the attention calculation and the evidence nodes for the key and value parts:
\begin{equation*}
\text{Attention}(Q, K, V) = \text{Softmax}\left(\frac{QK^T}{\sqrt{d_k}}\right)V
\end{equation*}
where $Q$ is the query projection of the claim graph and $K$ and $V$ are key and value projections of the evidence graph.

After cross-attention, we apply global mean pooling to get the average node feature to represent the graphs in the batch. For each batch, we calculate:
\[
\mathbf{g_X} = \frac{1}{|\mathcal{V}_X|} \sum_{v \in \mathcal{V}_X} h_X(v)
\]
where $X \in \{\text{evidence}, \text{claim}, \text{attended}\}$, and $h_X(v)$ is the node feature for $v$. We calculate the global mean for all three types: evidence graph, claim graphs, and cross-attention. We then concatenate them to get a combined representation of each sample:
\[
\mathbf{f} = \mathbf{g_{\text{evidence}}} \parallel \mathbf{g_{\text{claim}}} \parallel \mathbf{g_{\text{parallel}}}
\]
The combined features are fed into a classifier layer that makes the decision, outputting a score between 0 and 1 that represents how well the evidence supports the claim:
\[s = \sigma(W \cdot \mathbf{f} + b)\]
where $\sigma$ is the sigmoid activation function, $W$ is the weight matrix, $b$ is the bias term and $s \in [0, 1]$ represents the claim veracity score.

\begin{table*}[t!]
    \centering
    \begin{tabular}{lcccccc}
        \toprule
        & \multicolumn{2}{c}{\texttt{EVAL\_ALL} ($n=1145$)} & \multicolumn{2}{c}{\texttt{EVAL\_SUFFICIENT}} & \multicolumn{2}{c}{\texttt{EVAL\_COMMON} ($n=461$)} \\
        \cline{2-7}
        \textbf{Model} & \textbf{Accuracy} & \textbf{F1} & \textbf{Accuracy} & \textbf{F1} & \textbf{Accuracy} & \textbf{F1} \\
        \midrule
        Sonnet 3.7 & 0.6182 & 0.4969 & 0.8515 & 0.8325 & 0.8676 & 0.8571 \\
        Haiku 3.5 & 0.6692 & 0.5830 & \underline{0.8695} & \underline{0.8695} & {0.8915} & {0.8826} \\
        GPT 4o & \underline{0.6872} & \underline{0.6209} & \textbf{0.8828} & \textbf{0.8841} & \underline{0.9023} & \underline{0.8936}\\
        GPT 4o-mini & 0.6419 & 0.5438 & 0.8112 & 0.7891 & 0.8741 & 0.8619 \\
        EGMMG (ours) & \textbf{0.8305} & \textbf{0.8455} & - & - & \textbf{0.9305} & \textbf{0.9219} \\
        \bottomrule
    \end{tabular}
    \caption{Performance metrics comparison across different test sets. \texttt{EVAL\_SUFFICIENT} counts are different for each model.}
    \label{tab:model-comparison}
\end{table*}

\section{Experimental Setup}
Our work focuses on the image contextualization task and generates graph data based on image-caption datasets using reverse image search. Since the graph data generated for our task is effectively new, with different number of samples (depending on availability of online evidence) than the original (Table \ref{tab:dataset_statistics}), we compare the performance of our baseline method with frontier LLMs available at the time of writing: Claude Sonnet 3.7, Claude Haiku 3.5, GPT 4o, GPT 4o-mini. We focus our experiments and evaluation on the Factify dataset first, then discuss about robustness to other datasets and generalizability in subsection \S\ref{sec:generalizability}.

\subsection{Evaluation Sets}
\label{eval-sets}
For the Factify evaluation set, we use the validation set introduced in the Factify paper. The Facitify validation set consists of 7000 samples (1500 for each label type). Because we focus only on the postitive and negative labels, we extract 3000 samples (1500 each for \texttt{Refute} and \texttt{Support\_Multimodal} classes). We apply the evidence retrieval pipeline described in \S\ref{sec:evidence-retrieval}. At the end we have 1145 samples in the evaluation set.

We prompt the LLMs with the evidence document and the claim as input, and the LLMs are asked whether the evidence-claim text pairs are misinformation or not. (Prompts can be found in Appendix \ref{appendix-prompts} and Figure \ref{fig:fact-checking-prompt}.) The models abstain on some samples, and so we construct three evaluation sets based on abstentions.
\begin{enumerate}
    \item \texttt{EVAL\_ALL} ($n = 1145$): We prompt the models to answer regardless if they consider the evidence insufficient. (All models answer for all 1145 samples under this setting, except Sonnet which abstained on 64 samples even when prompted to answer strictly between ``True'' or ``False''.)
    \item \texttt{EVAL\_SUFFICIENT:} We prompt the models allowing them to abstain on samples that they do not consider answerable with the provided evidence. This has the obvious issue of the models choosing to only answer for samples that are “easy” for them to decide, abstaining from difficult ones. Our method does not abstain, but we include these results for completeness.
    \item \texttt{EVAL\_COMMON} ($n = 461$): One limitation with \texttt{EVAL\_SUFFICIENT} is that each of the models may not have the exact same samples, thus making comparison difficult. So we also take an overlap set that all the models consider answerable and have answered. 
    The \texttt{EVAL\_COMMON} set is given by
   \begin{equation*}
    \mathbf{ES_{Sonnet}} \cap \mathbf{ES_{GPT4oMini}} \cap \mathbf{ES_{Haiku}}
   \end{equation*}
   where $\mathbf{ES}$ is \texttt{EVAL\_SUFFICIENT}.
\end{enumerate}

\subsection{Generalization and Robustness}
\label{sec:generalizability}
In addition to evaluating the model on the Factify benchmark, we evaluate the generalizability of our pipeline on other datasets. For these tests, we first obtain online evidence and prepare graph data. We then perform train-test performance analysis on individual datasets. 

We focus on: 
\begin{itemize}[nosep]
    \item evaluating the model on different datasets using a standard 85:15 train-test split, training and evaluating the accuracy of the model on the test set,
    \item architecture ablation on a particular test set to understand the model's robustness.
\end{itemize}

\section{Results}
Table \ref{tab:model-comparison} presents a performance comparison between multiple LLMs and our proposed approach. The evaluation was conducted across the three test sets described in section \ref{eval-sets}. For each evaluation category, we report accuracy and F1 scores.

Our method outperforms the LLMs on both \texttt{EVAL\_ALL} and \texttt{EVAL\_COMMON} sets. As discussed earlier, our model does not abstain, so we do not have an \texttt{EVAL\_SUFFICIENT} set. On the \texttt{EVAL\_ALL} test set, EGMMG achieves an accuracy of $0.8305$ and F1 score of $0.8455$, substantially higher than GPT-4o ($0.6872/0.6209$), Haiku 3.5 ($0.6692/0.5830$), GPT4o-mini ($0.6419/0.5438$) and Sonnet 3.7 ($0.6182/0.4969$). 

On the \texttt{EVAL\_COMMON} set, EGMMG has $0.9305$ accuracy and $0.9219$ F1 scores, compared to GPT-4o's $0.9023/0.8936$. GPT-4o performs marginally better on \texttt{EVAL\_SUFFICIENT} ($0.8828/0.8841$) compared to Haiku 3.5 ($0.8695/0.8695$).

Table \ref{tab:dataset_comparison} presents our method's performance on the different datasets we have discussed. We report these numbers to discuss the generalization abilities of our baseline method. The best performance is achieved on the Factify dataset (85:15 train-test split) ($0.8248$). It performs relatively well on the COSMOS Test dataset ($0.7750$) as well. But the method struggles with the MMFB Val ($0.7100$) and MMFB Test ($0.6823$) sets. For a baseline approach, with no special modifications for individual datasets, the model maintains reasonable generalization capabilities.

\begin{table}[htbp]
\centering
\renewcommand{\arraystretch}{1.2}
\begin{tabular}{lcccc}
\toprule
\textbf{Dataset} & \textbf{Acc} \\
\hline
Factify & 0.8248 \\ 
MMFB Val & 0.7100 \\
MMFB Test & 0.6823 \\
COSMOS Test & 0.7750 \\ 
\bottomrule
\end{tabular}
\caption{Performance of our model on different datasets on 85:15 train-test split. For each dataset, we train on the train split and report performance on the unseen test split.}
\label{tab:dataset_comparison}
\end{table}

\section{Discussion}
Earlier work in image contextualization has focused on using metadata \cite{tonglet2024imagetellstorypredicting, tonglet2025covecontextveracityprediction}, image entities extraction \cite{aneja2023cosmos, ma-etal-2024-event} and LLM knowledge \cite{qi2024sniffermultimodallargelanguage, tahmasebi2024multimodalmisinformationdetectionusing}, among others. This work focuses on using related online text content only. Below we briefly discuss the performance of the model, its robustness to ablation and datasets and efficiency.

\subsection{Model Performance and Robustness}
Table \ref{tab:model-comparison} shows that our classifier performs better than frontier LLMs on the Factify evaluation sets. All methods have access to the same amount of data (in text or graph format).

To investigate the model's robustness and the contribution of individual components, we conduct an ablation study (Table \ref{tab:ablation-study}). We report the performances on the \texttt{EVAL\_COMMON} set. 

The full model achieves the best performance with $0.9305$ accuracy and $0.9219$ F1 score. Adding edge features (\S\ref{sec:edge-features}) causes a performance drop to $0.9132$ accuracy and $0.9$ F1 score, probably indicating that edge information adds noise and might need to be processed differently. A more substantial degradation occurs when using unweighted node embeddings (i.e. without a node feature projector and weight coefficients) ($0.8741/0.8473$) or reduced-dimension 384-dim node embeddings (instead of BERT's 768-dim) ($0.8872/0.8725$). 

Weighing the contributions of node label embeddings and node neighborhood information seems particularly important, as evidenced by the differences in accuracy and F1 score (especially F1 score). This is due to the 5 dimensional neighborhood structure information in the unweighted setup.

While further improvements seem possible by increasing the node dimensions and dedicated processing of edge features, we can see (from Tables \ref{tab:dataset_comparison} and \ref{tab:ablation-study}) that our method is robust.

\subsection{Efficiency}
The results are also encouraging from an efficiency point of view. Our model is significantly smaller compared to the LLMs we compare it against. The 768-dim node embeddings variant of our model has 10M parameters ($10,724,391$) and takes up around 41 megabytes of disk space. 

Due to the size and intermediate dimensions, the computational costs are also sizably small. We run our training and tests on a single NVIDIA T4. The inference is similarly cheap.

\begin{table}[t!]
    \centering
    \begin{tabular}{>{\raggedright\arraybackslash}p{0.45\columnwidth}|>{\centering\arraybackslash}p{0.2\columnwidth}|>{\centering\arraybackslash}p{0.2\columnwidth}}
        
        \toprule
        \textbf{Model Configuration} & \textbf{Accuracy} & \textbf{F1 Score} \\
        \midrule
        Full Model & 0.9305 & 0.9219 \\
        \hline
        + edge features & 0.9132 & 0.9 \\
        \hline
        unweighted node embeddings & 0.8741 & 0.8473 \\
        \hline
        384-dim node embeddings & 0.8872 & 0.8725 \\
        
\bottomrule
    \end{tabular}
    \caption{Ablation study results. Each row represents the performance when a specific component is removed or modified.}
    \label{tab:ablation-study}
\end{table}

\section{Conclusion}
In this work, we introduced a graph-based method to tackle the image contextualization task for multimodal out-of-context misinformation detection. We developed an online evidence retrieval pipeline and graph data generation method to ground images with textual evidence available online. Then we introduced a GNN-based method to learn misinformation classification over the generated graph data. We experimented with several publicly available datasets using our method. Our results show that using relevant text information, in the form of entity-relation graphs, is greatly effective in misinformation detection, evidenced by the performance of our proposed method over frontier LLMs when provided the same information. The effectiveness of the method also highlights possible improvements.

\section*{Limitations}
Some limitations of this work are highlighted below. 

First, our methods do not utilize the images directly. We use reverse search on images to get web pages with matches, but we do not process the image itself for information. Extracting actors and events from the image could potentially improve the model.

Second, the evidence retrieval method can be made more robust. Currently, images that do not have web page matches are discarded. Better (or multiple) image search methods could help improve web page retrieval. We currently also do not have a method to establish the relevance of evidences collected, thus depending completely on the image match and the cosine similarity between the evidence texts and the image embedding.

Similarly, there are aspects of the text-to-graph pipeline that have potential room for improvement. With more rules for entity and relations extraction, the method could extract more information relevant to the veracity detection task.

Our approach currently does not use existing large knowledge graphs (for example, ConceptNet) to help incorporate real-world logic. It did not fit the current research scope, but might assist the task with common-sense knowledge.

Earlier work has been about using metadata only and this work focuses on using related text content only. Combining these methods for evidence and adding LLMs in the workflow would be a worth-exploring direction.


\bibliography{custom}
\bibliographystyle{acl_natbib}

\clearpage

\appendix

\begin{figure*}[t!]
    \centering
    \includegraphics[width=\linewidth]{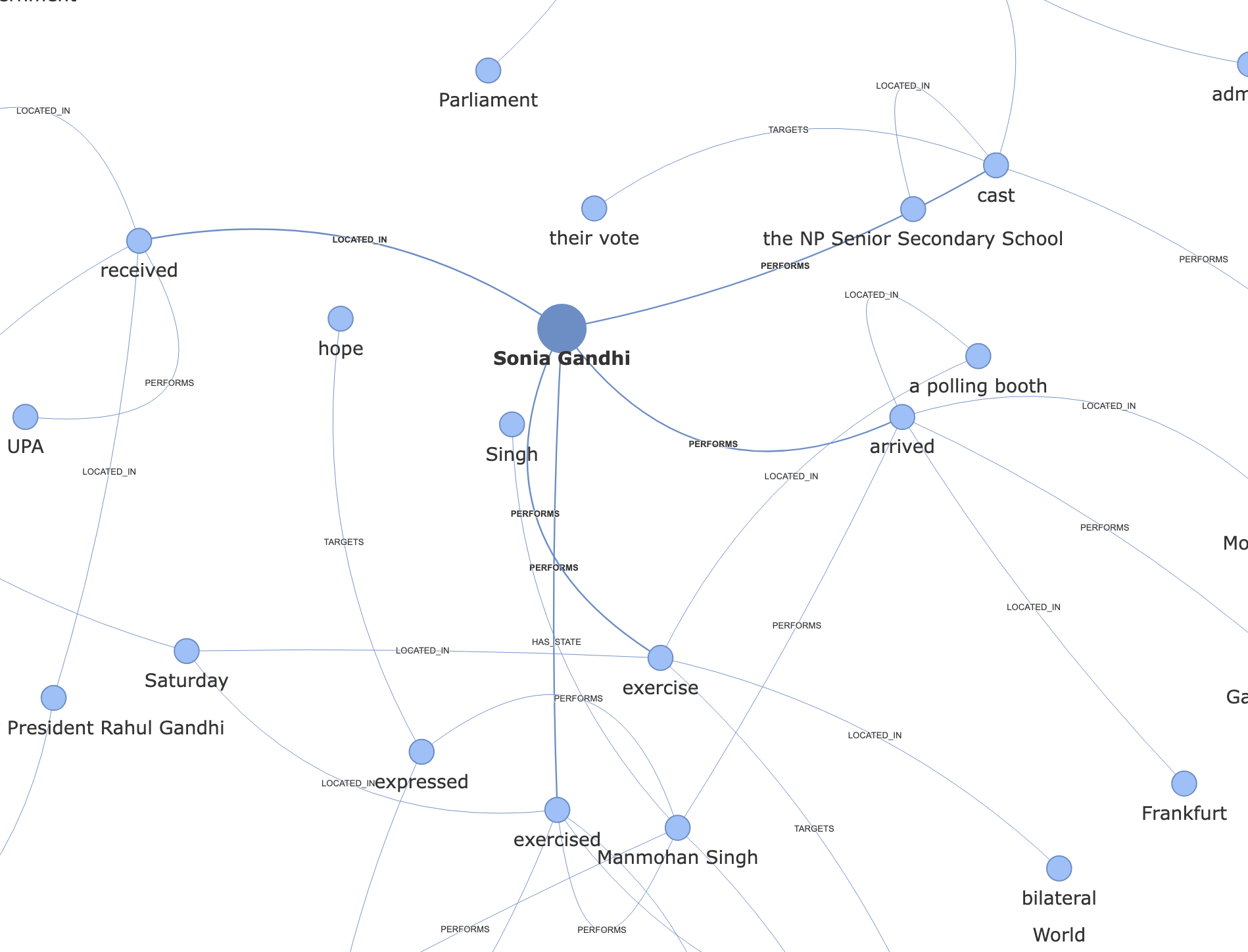}
    \caption{Evidence graph generated by EGMMG for the example in Figure \ref{fig:gandhi-voting-comprehensive}}
    \label{fig:gandhi-voting-comprehensive-evidence}
\end{figure*}

\section{Model Details}
Our baseline model has 2 convolution layers for message passing between the nodes. For node label embeddings, we experiment with two language models: a) \texttt{BERT-Base} (768 embedding size) and b) SentenceTransformer's \texttt{all-MiniLM-L6-v2} (384 embedding size). The node embeddings for our model includes the label embedding and the neighborhood structure information. To accomplish this we use a feature projector to map text embeddings (384- or 768-dim) and structural features (5-dim) to a common (389-dim or 773-dim) space.

We use TransformerConv as our convolution layer after a set of experiments (see Table \ref{tab:conv_type_performance_merged}. We also use multiheaded attention. Among the two conv layers, the first layer uses 4 attention heads and the second layer uses 2 attention heads.

The hidden dimension of the model is 1024.

All processing and experiments (graph data generation, training, inference) were run on an NVIDIA T4.

\begin{table}[!ht]
\centering
\renewcommand{\arraystretch}{1.2}
\begin{tabular}{lc}
\toprule
\textbf{Hyperparameter} & \\
\hline
Node label embedding & 768 \\
Hidden dimensions & 1024 \\
Conv. layers & 2 \\ 
Learning rate & 3e-4 \\
Batch size & 64 \\
No of parameters & 10,724,391 \\
\bottomrule
\end{tabular}
\caption{Model details.}
\label{tab:model_architecture}
\end{table}

\section{Knowledge Graph Construction}
\label{graph-construction}
Here we describe the rules for graph construction for our system, including node and edge type taxonomies, their extraction and relationship formation.

\subsection{Entities as nodes}
First let's define what object types we consider a node: 
\texttt{ENTITY}, \texttt{EVENT}, \texttt{STATE}, \texttt{LOCATION}, \texttt{TIME} and \texttt{ATTRIBUTE}.
\\\\
\textbf{Entity Type Assignment} \, Entities are classified based on their NER labels (when present) into one of the aforementioned subtypes. When the label is not present or doesn't match any predefined subtype, we default to the \texttt{ENTITY} type.
\\\\
\textbf{Node Identification and Deduplication} \, We map entity label to node IDs using the original- and lowercase variants. For entities not beginning with "the", we additionally map "the [entity]" to the same node ID to handle different references to the same entity to ensure consistency.

\subsection{Relations as edges}
As mentioned earlier in the document, the edge types we focus on are: \texttt{PERFORMS}, \texttt{EXPERIENCES}, \texttt{TARGETS}, \texttt{LOCATED\_IN}, \texttt{HAS\_STATE} and \texttt{SAME\_AS}.
\\\\
\textbf{Verbs as events} \, For tokens with \texttt{VERB} part-of-speech, we create \texttt{EVENT} nodes. We establish different relationship types based on following rules:
\begin{itemize}[nosep]
    \item \texttt{nsubj} (subject) creates a \texttt{PERFORMS} edge from subject to verb,
    \item \texttt{nsubjpass} (passive subject) creates an \texttt{EXPERIENCES} edge from verb to subject,
    \item \texttt{dobj} or \texttt{pobj} (direct/prepositional objects) creates \texttt{TARGETS} edges from verb to object.
\end{itemize}\leavevmode
\\
\textbf{Prepositions for locations} \, When a verb has a child with \texttt{prep} dependency and text ``in'', ``at'', or ``on'', we create a \texttt{LOCATED\_IN} edge from the verb to location. Similarly, for tokens with \texttt{prep} dependency and text ``in'' where both head and child exist in the node map, we create a \texttt{LOCATED\_IN} edge from head to child.
\\\\
\textbf{Attribute and \texttt{SAME\_AS} edges} \, If a compound and its head are in the node map and we find the compound phrase exists, we create \texttt{HAS\_STATE} edge from the compound to the head entity. And, the \texttt{SAME\_AS} edge is used for co-references. This is most likely handled by the node deduplication step.

\section{Prompts}
\label{appendix-prompts}
Since we do not have direct baselines to compare our method against, we use LLMs on the eval sets we prepare (Table \ref{tab:model-comparison}).

The prompt we use for the LLMs is provided in Figure \ref{fig:fact-checking-prompt}.

For the \texttt{EVAL\_SUFFICIENT} set, we edit the prompt to allow the model to choose among three answers: ``true'', ``false'' and ``not enough information''.

\section{Data Licenses}
\begin{itemize}
    \item \textbf{Factify:} CC BY 4.0
    \item \textbf{MMFakeBench:} CC BY-NC-SA 4.0
    \item \textbf{COSMOS:} Academic research only
\end{itemize}

All data provided under this work is licensed under CC BY-NC-SA 4.0 because the data we originally use are released under this license.

\begin{figure*}[h!]
\fbox{\begin{minipage}{\textwidth}
\texttt{You are a fact-checking assistant tasked with evaluating the accuracy of a claim based on evidence provided. Your goal is to determine whether the claim is true or false based solely on the evidence. Do not consider external knowledge or information not included in the evidence.\\\\
Instructions:\\
1. Carefully read the evidence document, which consists of excerpts from multiple news articles.\\
2. Analyze the claim provided and compare it to the evidence.\\
3. Respond with "true" or "false" based on your analysis. Do not provide explanations or additional commentary.\\
\\
EVIDENCE: \{evidence\}\\
\\
CLAIM: \{claim\}\\
\\
Your response should be exactly one of: TRUE, FALSE.\\
\\
YOUR RESPONSE:}
\end{minipage}}
\caption{Prompt used to evaluate misinformation detection performance of LLMs (Sonnet, Haiku, GPT). For the \texttt{EVAL\_SUFFICIENT} set, we allow one more option: ``not enough information''.}
\label{fig:fact-checking-prompt}
\end{figure*}

\begin{table*}[!ht]
\centering
\renewcommand{\arraystretch}{1.2}
\begin{tabular}{lccccccc}
\toprule
 & \multicolumn{4}{c}{\textbf{384-dim embeddings}} & \multicolumn{3}{c}{\textbf{768-dim embeddings}} \\
\cmidrule(lr){2-5} \cmidrule(lr){6-8}
\textbf{Conv Type} & \textbf{Run 1} & \textbf{Run 2} & \textbf{Run 3} & \textbf{Run 4} & \textbf{Run 1} & \textbf{Run 2} & \textbf{Run 3} \\
\midrule
GatConv        & 0.8059 & 0.8140 & 0.8086 & 0.8181 & \textbf{0.8288} & 0.8235 & 0.8221 \\
Gatv2Conv      & 0.8099 & 0.8207 & 0.8194 & 0.8248 & 0.8221 & 0.8221 & 0.8194 \\
TransformerConv & 0.8180 & 0.8005 & \textbf{0.8315} & 0.8221 & 0.8221 & 0.8181 & 0.8248 \\
\bottomrule
\end{tabular}
\caption{Performance comparison of graph convolution methods across multiple runs for 384-dim and 768-dim node label embeddings on the Factify 85:15 split. Best performance per embedding dimension is in \textbf{bold}.}
\label{tab:conv_type_performance_merged}
\end{table*}

\end{document}

%% file: relatedworks.tex
\section{Related work}
\subsection{Attention-based GNNs}
\citet{veličković2018graphattentionnetworks} introduced graph attention networks (GAT) that used attention, popular in sequence-based natural language tasks, to tackle irregular graph structures, which GCN \cite{kipf2017semisupervisedclassificationgraphconvolutional} before them did not handle effectively. 

Other than the standard GAT and GCN, there have been approaches like Graph Transformer \cite{shi2021maskedlabelpredictionunified}, which uses node features and labels jointly, and GATv2 \cite{brody2022attentivegraphattentionnetworks}, which calculates dynamic attention as opposed to GAT's static.

GNN (attention) methods are now being specialized for particular tasks like drug discovery, material property prediction, misinformation detection, etc \cite{attention-drug, lu-li-2020-gcan}. More work in refining and making them more adoptable via scalability and interpretability is underway \cite{kazi2021iagcninterpretableattentionbased}.

\subsection{OOC Misinformation Detection}
Early methods of detecting OOC misinformation focused on image-text similarity and object alignment \cite{aneja2023cosmos} and researchers designed various LLM-based architectures for it \cite{qi2024sniffermultimodallargelanguage, aneja2023cosmos, xuan2024lemmalvlmenhancedmultimodalmisinformation, tahmasebi2024multimodalmisinformationdetectionusing}. However, these models were limited in the information available to them. Architectures were proposed to mitigate this limitation and to incorporate external information from the internet \cite{abdelnabi2022open}. For example, \citet{abdelnabi2022open} suggested gathering external knowledge regarding both the image and the caption of the (image, caption) pair to detect OOC misinformation. \citet{papadopoulos2025} showed that providing more context by adding external sources improved performance, even with relatively simple models. Our work builds on this concept as we also focus on providing external context to image-caption pairs. 

\citet{qi2024sniffermultimodallargelanguage} proposed the SNIFFER model, which not only detects OOC misinformation, but also provides an explanation for the model's choice, thus improving the interpretability of the model. \citet{tonglet2024imagetellstorypredicting} suggested that providing context to images by asking various questions through an LLM pipeline could establish the factuality of a sample. However, LLM-based models are resource-intensive to train and have the potential to hallucinate. Graph-based approaches make the systems more accurate and explainable using causal methods \cite{opsahl2024factfictionimprovingfact, wangshu23explainable, tan-etal-2024-enhancing-fact, lu-li-2020-gcan}.

\subsection{Datasets}
\label{sec:datasets}
Misinformation detection on text is well-studied, with a great amount of work. Research in multimodal misinformation detection is also picking up, due to growing necessity and interest. As a result, there are several datasets, distant-supervised and manually annotated. There are also fine-grained divisions along which these datasets are categorized: textual distortion, visual distortion, edited image, repurposed image, etc. FEVER \cite{thorne-etal-2018-fever} and Politifact \cite{shu2019fakenewsnetdatarepositorynews} focus on textual distortion, especially rumors. Fakeddit \cite{nakamura-etal-2020-fakeddit} was collected from over 1 million samples and included various categories of fake news, distantly supervised. Factify is another multimodal fact verification dataset, collected from tweets of US and Indian news sources \cite{Mishra2022FACTIFYAM}. NewsCLIPpings \cite{luo-etal-2021-newsclippings} and COSMOS \cite{aneja2023cosmos} focused particularly on OOC misinformation too. More recently, especially to tackle the issues related to distortion using AI (textual, visually altered, generated), LLMFake \cite{chen2024llmgenerated} and MMFakeBench \cite{liu2024mmfakebench} have been introduced.

%% file: acl2023.bbl
\begin{thebibliography}{41}
\expandafter\ifx\csname natexlab\endcsname\relax\def\natexlab#1{#1}\fi

\bibitem[{Abdelnabi et~al.(2022)Abdelnabi, Hasan, and Fritz}]{abdelnabi2022open}
Sahar Abdelnabi, Rakibul Hasan, and Mario Fritz. 2022.
\newblock Open-domain, content-based, multi-modal fact-checking of out-of-context images via online resources.
\newblock In \emph{Proceedings of the IEEE/CVF conference on computer vision and pattern recognition}, pages 14940--14949.

\bibitem[{Aneja et~al.(2023)Aneja, Bregler, and Nie{\ss}ner}]{aneja2023cosmos}
Shivangi Aneja, Chris Bregler, and Matthias Nie{\ss}ner. 2023.
\newblock Cosmos: catching out-of-context image misuse using self-supervised learning.
\newblock In \emph{Proceedings of the AAAI conference on artificial intelligence}, volume~37, pages 14084--14092.

\bibitem[{Brody et~al.(2022)Brody, Alon, and Yahav}]{brody2022attentivegraphattentionnetworks}
Shaked Brody, Uri Alon, and Eran Yahav. 2022.
\newblock \href {http://arxiv.org/abs/2105.14491} {How attentive are graph attention networks?}

\bibitem[{Brown et~al.(2020)Brown, Mann, Ryder, Subbiah, Kaplan, Dhariwal, Neelakantan, Shyam, Sastry, Askell, Agarwal, Herbert-Voss, Krueger, Henighan, Child, Ramesh, Ziegler, Wu, Winter, Hesse, Chen, Sigler, Litwin, Gray, Chess, Clark, Berner, McCandlish, Radford, Sutskever, and Amodei}]{brown2020languagemodelsfewshotlearners}
Tom~B. Brown, Benjamin Mann, Nick Ryder, Melanie Subbiah, Jared Kaplan, Prafulla Dhariwal, Arvind Neelakantan, Pranav Shyam, Girish Sastry, Amanda Askell, Sandhini Agarwal, Ariel Herbert-Voss, Gretchen Krueger, Tom Henighan, Rewon Child, Aditya Ramesh, Daniel~M. Ziegler, Jeffrey Wu, Clemens Winter, Christopher Hesse, Mark Chen, Eric Sigler, Mateusz Litwin, Scott Gray, Benjamin Chess, Jack Clark, Christopher Berner, Sam McCandlish, Alec Radford, Ilya Sutskever, and Dario Amodei. 2020.
\newblock \href {http://arxiv.org/abs/2005.14165} {Language models are few-shot learners}.

\bibitem[{Chen and Shu(2024)}]{chen2024llmgenerated}
Canyu Chen and Kai Shu. 2024.
\newblock \href {https://openreview.net/forum?id=ccxD4mtkTU} {Can {LLM}-generated misinformation be detected?}
\newblock In \emph{The Twelfth International Conference on Learning Representations}.

\bibitem[{Denniss and Lindberg(2025)}]{social-media-and-misinformation}
Emily Denniss and Rebecca Lindberg. 2025.
\newblock \href {https://doi.org/10.1093/heapro/daaf023} {Social media and the spread of misinformation: infectious and a threat to public health}.
\newblock \emph{Health Promotion International}, 40(2):daaf023.

\bibitem[{Devlin et~al.(2019)Devlin, Chang, Lee, and Toutanova}]{devlin-etal-2019-bert}
Jacob Devlin, Ming-Wei Chang, Kenton Lee, and Kristina Toutanova. 2019.
\newblock \href {https://doi.org/10.18653/v1/N19-1423} {{BERT}: Pre-training of deep bidirectional transformers for language understanding}.
\newblock In \emph{Proceedings of the 2019 Conference of the North {A}merican Chapter of the Association for Computational Linguistics: Human Language Technologies, Volume 1 (Long and Short Papers)}, pages 4171--4186, Minneapolis, Minnesota. Association for Computational Linguistics.

\bibitem[{Dufour et~al.(2024)Dufour, Pathak, Samangouei, Hariri, Deshetti, Dudfield, Guess, Escayola, Tran, Babakar, and Bregler}]{dufour2024ammebalargescalesurveydataset}
Nicholas Dufour, Arkanath Pathak, Pouya Samangouei, Nikki Hariri, Shashi Deshetti, Andrew Dudfield, Christopher Guess, Pablo~Hernández Escayola, Bobby Tran, Mevan Babakar, and Christoph Bregler. 2024.
\newblock \href {http://arxiv.org/abs/2405.11697} {Ammeba: A large-scale survey and dataset of media-based misinformation in-the-wild}.

\bibitem[{Fazio(2020)}]{fazio2020out}
Lisa Fazio. 2020.
\newblock Out-of-context photos are a powerful low-tech form of misinformation.
\newblock \emph{The Conversation}, 14(1).

\bibitem[{Fisher et~al.(2016)Fisher, Cox, and Hermann}]{wapo-pizzagate}
Marc Fisher, John~Woodrow Cox, and Peter Hermann. 2016.
\newblock \href {https://www.washingtonpost.com/local/pizzagate-from-rumor-to-hashtag-to-gunfire-in-dc/2016/12/06/4c7def50-bbd4-11e6-94ac-3d324840106c_story.html} {Pizzagate: From rumor, to hashtag, to gunfire in d.c.}

\bibitem[{Gao et~al.(2024)Gao, Xiong, Gao, Jia, Pan, Bi, Dai, Sun, Wang, and Wang}]{gao2024retrievalaugmentedgenerationlargelanguage}
Yunfan Gao, Yun Xiong, Xinyu Gao, Kangxiang Jia, Jinliu Pan, Yuxi Bi, Yi~Dai, Jiawei Sun, Meng Wang, and Haofen Wang. 2024.
\newblock \href {http://arxiv.org/abs/2312.10997} {Retrieval-augmented generation for large language models: A survey}.

\bibitem[{Honnibal et~al.(2020)Honnibal, Montani, Van~Landeghem, and Boyd}]{Honnibal_spaCy_Industrial-strength_Natural_2020}
Matthew Honnibal, Ines Montani, Sofie Van~Landeghem, and Adriane Boyd. 2020.
\newblock \href {https://doi.org/10.5281/zenodo.1212303} {{spaCy: Industrial-strength Natural Language Processing in Python}}.

\bibitem[{Islam et~al.(2020)Islam, Islam, Sarkar, Khan, Kamal, Hasan, Kabir, Yeasmin, Islam, Chowdhury, Anwar, Chughtai, and Seale}]{Islam2020COVID19RelatedIA}
S~M~Hasibul Islam, S~M~Hasibul Islam, Tonmoy Sarkar, Sazzad~Hossain Khan, Abu Hena~Mostafa Kamal, S.~M.~Murshid Hasan, Alamgir Kabir, Dalia Yeasmin, Mohammad~A. Islam, Kamal Ibne~Amin Chowdhury, Kazi~Selim Anwar, Abrar~Ahmad Chughtai, and Holly Seale. 2020.
\newblock \href {https://api.semanticscholar.org/CorpusID:221121856} {Covid-19–related infodemic and its impact on public health: A global social media analysis}.
\newblock \emph{The American Journal of Tropical Medicine and Hygiene}, 103:1621 -- 1629.

\bibitem[{Kazi et~al.(2021)Kazi, Farghadani, and Navab}]{kazi2021iagcninterpretableattentionbased}
Anees Kazi, Soroush Farghadani, and Nassir Navab. 2021.
\newblock \href {http://arxiv.org/abs/2103.15587} {Ia-gcn: Interpretable attention based graph convolutional network for disease prediction}.

\bibitem[{Kim et~al.(2023)Kim, Park, Kwon, Jo, Thorne, and Choi}]{kim2023factkgfactverificationreasoning}
Jiho Kim, Sungjin Park, Yeonsu Kwon, Yohan Jo, James Thorne, and Edward Choi. 2023.
\newblock \href {http://arxiv.org/abs/2305.06590} {Factkg: Fact verification via reasoning on knowledge graphs}.

\bibitem[{Kipf and Welling(2017)}]{kipf2017semisupervisedclassificationgraphconvolutional}
Thomas~N. Kipf and Max Welling. 2017.
\newblock \href {http://arxiv.org/abs/1609.02907} {Semi-supervised classification with graph convolutional networks}.

\bibitem[{Liu et~al.(2024)Liu, Li, Li, Xia, Cui, Huang, Huang, Deng, and He}]{liu2024mmfakebench}
Xuannan Liu, Zekun Li, Peipei Li, Shuhan Xia, Xing Cui, Linzhi Huang, Huaibo Huang, Weihong Deng, and Zhaofeng He. 2024.
\newblock Mmfakebench: A mixed-source multimodal misinformation detection benchmark for lvlms.
\newblock \emph{arXiv preprint arXiv:2406.08772}.

\bibitem[{Lu and Li(2020)}]{lu-li-2020-gcan}
Yi-Ju Lu and Cheng-Te Li. 2020.
\newblock \href {https://doi.org/10.18653/v1/2020.acl-main.48} {{GCAN}: Graph-aware co-attention networks for explainable fake news detection on social media}.
\newblock In \emph{Proceedings of the 58th Annual Meeting of the Association for Computational Linguistics}, pages 505--514, Online. Association for Computational Linguistics.

\bibitem[{Luo et~al.(2021)Luo, Darrell, and Rohrbach}]{luo-etal-2021-newsclippings}
Grace Luo, Trevor Darrell, and Anna Rohrbach. 2021.
\newblock \href {https://doi.org/10.18653/v1/2021.emnlp-main.545} {{N}ews{CLIP}pings: {A}utomatic {G}eneration of {O}ut-of-{C}ontext {M}ultimodal {M}edia}.
\newblock In \emph{Proceedings of the 2021 Conference on Empirical Methods in Natural Language Processing}, pages 6801--6817, Online and Punta Cana, Dominican Republic. Association for Computational Linguistics.

\bibitem[{Ma et~al.(2024)Ma, Luo, Guo, Zeng, Hao, and Zhao}]{ma-etal-2024-event}
Zihan Ma, Minnan Luo, Hao Guo, Zhi Zeng, Yiran Hao, and Xiang Zhao. 2024.
\newblock \href {https://doi.org/10.18653/v1/2024.acl-long.316} {Event-radar: Event-driven multi-view learning for multimodal fake news detection}.
\newblock In \emph{Proceedings of the 62nd Annual Meeting of the Association for Computational Linguistics (Volume 1: Long Papers)}, pages 5809--5821, Bangkok, Thailand. Association for Computational Linguistics.

\bibitem[{Maynez et~al.(2020)Maynez, Narayan, Bohnet, and McDonald}]{maynez-etal-2020-faithfulness}
Joshua Maynez, Shashi Narayan, Bernd Bohnet, and Ryan McDonald. 2020.
\newblock \href {https://doi.org/10.18653/v1/2020.acl-main.173} {On faithfulness and factuality in abstractive summarization}.
\newblock In \emph{Proceedings of the 58th Annual Meeting of the Association for Computational Linguistics}, pages 1906--1919, Online. Association for Computational Linguistics.

\bibitem[{Mishra et~al.(2022)Mishra, Suryavardan, Bhaskar, Chopra, Reganti, Patwa, Das, Chakraborty, Sheth, and Ekbal}]{Mishra2022FACTIFYAM}
Shreyash Mishra, S~Suryavardan, Amrit Bhaskar, Parul Chopra, Aishwarya~N. Reganti, Parth Patwa, Amitava Das, Tanmoy Chakraborty, A.~Sheth, and Asif Ekbal. 2022.
\newblock \href {https://api.semanticscholar.org/CorpusID:252016186} {Factify: A multi-modal fact verification dataset}.
\newblock In \emph{DE-FACTIFY@AAAI}.

\bibitem[{Nakamura et~al.(2020)Nakamura, Levy, and Wang}]{nakamura-etal-2020-fakeddit}
Kai Nakamura, Sharon Levy, and William~Yang Wang. 2020.
\newblock \href {https://aclanthology.org/2020.lrec-1.755/} {{F}akeddit: A new multimodal benchmark dataset for fine-grained fake news detection}.
\newblock In \emph{Proceedings of the Twelfth Language Resources and Evaluation Conference}, pages 6149--6157, Marseille, France. European Language Resources Association.

\bibitem[{Opsahl(2024)}]{opsahl2024factfictionimprovingfact}
Tobias~A. Opsahl. 2024.
\newblock \href {http://arxiv.org/abs/2408.07453} {Fact or fiction? improving fact verification with knowledge graphs through simplified subgraph retrievals}.

\bibitem[{Papadopoulos et~al.(2025)Papadopoulos, Koutlis, Papadopoulos, and Petrantonakis}]{papadopoulos2025}
Stefanos-Iordanis Papadopoulos, Christos Koutlis, Symeon Papadopoulos, and Panagiotis~C. Petrantonakis. 2025.
\newblock \href {https://doi.org/10.1109/WACV61041.2025.00544} {{ Similarity Over Factuality: Are we Making Progress on Multimodal Out-of-Context Misinformation Detection? }}.
\newblock In \emph{2025 IEEE/CVF Winter Conference on Applications of Computer Vision (WACV)}, pages 5041--5050, Los Alamitos, CA, USA. IEEE Computer Society.

\bibitem[{Qi et~al.(2024)Qi, Yan, Hsu, and Lee}]{qi2024sniffermultimodallargelanguage}
Peng Qi, Zehong Yan, Wynne Hsu, and Mong~Li Lee. 2024.
\newblock Sniffer: Multimodal large language model for explainable out-of-context misinformation detection.
\newblock In \emph{Proceedings of the IEEE/CVF Conference on Computer Vision and Pattern Recognition (CVPR)}, pages 13052--13062.

\bibitem[{Radford et~al.(2019)Radford, Wu, Child, Luan, Amodei, and Sutskever}]{Radford2019LanguageMA}
Alec Radford, Jeff Wu, Rewon Child, David Luan, Dario Amodei, and Ilya Sutskever. 2019.
\newblock \href {https://api.semanticscholar.org/CorpusID:160025533} {Language models are unsupervised multitask learners}.

\bibitem[{Reimers and Gurevych(2019)}]{reimers-2019-sentence-bert}
Nils Reimers and Iryna Gurevych. 2019.
\newblock \href {https://arxiv.org/abs/1908.10084} {Sentence-bert: Sentence embeddings using siamese bert-networks}.
\newblock In \emph{Proceedings of the 2019 Conference on Empirical Methods in Natural Language Processing}. Association for Computational Linguistics.

\bibitem[{Shi et~al.(2021)Shi, Huang, Feng, Zhong, Wang, and Sun}]{shi2021maskedlabelpredictionunified}
Yunsheng Shi, Zhengjie Huang, Shikun Feng, Hui Zhong, Wenjin Wang, and Yu~Sun. 2021.
\newblock \href {http://arxiv.org/abs/2009.03509} {Masked label prediction: Unified message passing model for semi-supervised classification}.

\bibitem[{Shu et~al.(2019)Shu, Mahudeswaran, Wang, Lee, and Liu}]{shu2019fakenewsnetdatarepositorynews}
Kai Shu, Deepak Mahudeswaran, Suhang Wang, Dongwon Lee, and Huan Liu. 2019.
\newblock \href {http://arxiv.org/abs/1809.01286} {Fakenewsnet: A data repository with news content, social context and spatialtemporal information for studying fake news on social media}.

\bibitem[{Shuster et~al.(2021)Shuster, Poff, Chen, Kiela, and Weston}]{shuster-etal-2021-retrieval-augmentation}
Kurt Shuster, Spencer Poff, Moya Chen, Douwe Kiela, and Jason Weston. 2021.
\newblock \href {https://doi.org/10.18653/v1/2021.findings-emnlp.320} {Retrieval augmentation reduces hallucination in conversation}.
\newblock In \emph{Findings of the Association for Computational Linguistics: EMNLP 2021}, pages 3784--3803, Punta Cana, Dominican Republic. Association for Computational Linguistics.

\bibitem[{Tahmasebi et~al.(2024)Tahmasebi, Müller-Budack, and Ewerth}]{tahmasebi2024multimodalmisinformationdetectionusing}
Sahar Tahmasebi, Eric Müller-Budack, and Ralph Ewerth. 2024.
\newblock \href {http://arxiv.org/abs/2407.14321} {Multimodal misinformation detection using large vision-language models}.

\bibitem[{Tan et~al.(2024)Tan, Desai, and Sengamedu}]{tan-etal-2024-enhancing-fact}
Fiona~Anting Tan, Jay Desai, and Srinivasan~H. Sengamedu. 2024.
\newblock \href {https://doi.org/10.18653/v1/2024.fever-1.20} {Enhancing fact verification with causal knowledge graphs and transformer-based retrieval for deductive reasoning}.
\newblock In \emph{Proceedings of the Seventh Fact Extraction and VERification Workshop (FEVER)}, pages 151--169, Miami, Florida, USA. Association for Computational Linguistics.

\bibitem[{Thorne et~al.(2018)Thorne, Vlachos, Christodoulopoulos, and Mittal}]{thorne-etal-2018-fever}
James Thorne, Andreas Vlachos, Christos Christodoulopoulos, and Arpit Mittal. 2018.
\newblock \href {https://doi.org/10.18653/v1/N18-1074} {{FEVER}: a large-scale dataset for fact extraction and {VER}ification}.
\newblock In \emph{Proceedings of the 2018 Conference of the North {A}merican Chapter of the Association for Computational Linguistics: Human Language Technologies, Volume 1 (Long Papers)}, pages 809--819, New Orleans, Louisiana. Association for Computational Linguistics.

\bibitem[{Tonglet et~al.(2024)Tonglet, Moens, and Gurevych}]{tonglet2024imagetellstorypredicting}
Jonathan Tonglet, Marie-Francine Moens, and Iryna Gurevych. 2024.
\newblock \href {http://arxiv.org/abs/2408.09939} {"image, tell me your story!" predicting the original meta-context of visual misinformation}.

\bibitem[{Tonglet et~al.(2025)Tonglet, Thiem, and Gurevych}]{tonglet2025covecontextveracityprediction}
Jonathan Tonglet, Gabriel Thiem, and Iryna Gurevych. 2025.
\newblock \href {http://arxiv.org/abs/2502.01194} {Cove: Context and veracity prediction for out-of-context images}.

\bibitem[{Urbani(2020)}]{urbani5pils}
Shaydanay Urbani. 2020.
\newblock Verifying online information.

\bibitem[{Veličković et~al.(2018)Veličković, Cucurull, Casanova, Romero, Liò, and Bengio}]{veličković2018graphattentionnetworks}
Petar Veličković, Guillem Cucurull, Arantxa Casanova, Adriana Romero, Pietro Liò, and Yoshua Bengio. 2018.
\newblock \href {http://arxiv.org/abs/1710.10903} {Graph attention networks}.

\bibitem[{Wang and Shu(2023)}]{wangshu23explainable}
Haoran Wang and Kai Shu. 2023.
\newblock \href {https://doi.org/10.18653/v1/2023.findings-emnlp.416} {Explainable claim verification via knowledge-grounded reasoning with large language models}.
\newblock In \emph{Findings of the Association for Computational Linguistics: EMNLP 2023}, pages 6288--6304, Singapore. Association for Computational Linguistics.

\bibitem[{Xuan et~al.(2024)Xuan, Yi, Yang, Wu, Fung, and Ji}]{xuan2024lemmalvlmenhancedmultimodalmisinformation}
Keyang Xuan, Li~Yi, Fan Yang, Ruochen Wu, Yi~R. Fung, and Heng Ji. 2024.
\newblock \href {http://arxiv.org/abs/2402.11943} {Lemma: Towards lvlm-enhanced multimodal misinformation detection with external knowledge augmentation}.

\bibitem[{Zhang et~al.(2024)Zhang, Liu, Liu, Liu, Lin, Huang, and Ning}]{attention-drug}
Yang Zhang, Caiqi Liu, Mujiexin Liu, Tianyuan Liu, Hao Lin, Cheng-Bing Huang, and Lin Ning. 2024.
\newblock \href {https://doi.org/10.1093/bib/bbad467} {Attention is all you need: utilizing attention in ai-enabled drug discovery}.
\newblock \emph{Briefings in Bioinformatics}, 25(1):bbad467.

\end{thebibliography}
